\documentclass[letterpaper, 10 pt, conference]{ieeeconf}  

\IEEEoverridecommandlockouts                              

\usepackage{amsmath}
\usepackage{amssymb}
\usepackage{booktabs}
\usepackage{multirow}
\newcommand{\boldhline}{\specialrule{0.15em}{0em}{0.1em}}

\usepackage{adjustbox}
\usepackage{colortbl,xcolor}
\usepackage{subcaption}
\usepackage[capitalize,sort&compress,nameinlink]{cleveref}

\crefname{section}{Sec.}{Secs.}
\Crefname{section}{Section}{Sections}
\Crefname{table}{Table}{Tables}
\crefname{table}{Tab.}{Tabs.}

\title{\LARGE \bf
SAILS: Segment Anything with Incrementally Learned Semantics for Task-Invariant and Training-Free Continual Learning
}

\author{
	Shishir Muralidhara$^1$ \qquad Didier Stricker$^{1,2}$ \qquad René Schuster$^{1,2}$  \\
	$^1$German Research Center for Artificial Intelligence (DFKI) \\
	$^2$RPTU -- University of Kaiserslautern-Landau, Kaiserslautern \\
	{\texttt{firstname.lastname@dfki.de}
     }
}

\begin{document}

\maketitle
\thispagestyle{empty}
\pagestyle{empty}

\begin{abstract}

Continual learning remains constrained by the need for repeated retraining, high computational costs, and the persistent challenge of forgetting.
These factors significantly limit the applicability of continual learning in real-world settings, as iterative model updates require significant computational resources and inherently exacerbate forgetting.
We present SAILS -- Segment Anything with Incrementally Learned Semantics, a training-free framework for Class-Incremental Semantic Segmentation (CISS) that sidesteps these challenges entirely.
SAILS leverages foundational models to decouple CISS into two stages: Zero-shot region extraction using Segment Anything Model (SAM),  followed by semantic association through prototypes in a fixed feature space. 
SAILS incorporates selective intra-class clustering, resulting in multiple prototypes per class to better model intra-class variability.
Our results demonstrate that, despite requiring no incremental training, SAILS typically surpasses the performance of existing training-based approaches on standard CISS datasets, particularly in long and challenging task sequences where forgetting tends to be most severe.
By avoiding parameter updates, SAILS completely eliminates forgetting and maintains consistent, task-invariant performance.
Furthermore, SAILS exhibits positive backward transfer, where the introduction of new classes can enhance performance on previous classes. \looseness-1

\end{abstract}

\section{Introduction}

Continual learning is a paradigm where models continuously adapt to new data and tasks over time, unlike traditional isolated learning \cite{isolated_learning} that rely on fixed datasets and predefined objectives.
However, this process introduces the challenge of catastrophic forgetting \cite{catastrophic_forgetting}, where learning new tasks overwrites previously learned knowledge, degrading performance on previous tasks.
Forgetting is an inherent consequence of the training process, as learning requires updating model parameters that adversely interferes with previously learned weights. 
The severity of forgetting largely depends on the task sequence, with longer sequences amplifying the cumulative effect of incremental updates and worsening forgetting over time.
The stability-plasticity dilemma \cite{stability_plasticity_dilemma} also influences forgetting, highlighting the challenge of retaining prior knowledge (stability) while adapting to new information (plasticity). 
While mitigating forgetting remains the primary challenge in continual learning, efficiency is an equally crucial factor, particularly for real-world applications.
Continual learning is inherently storage-efficient, as it works under the confines of restricted or no access to previous data.
However, computational efficiency has only recently received attention, despite its importance for practical deployment.
Most CL methods assume access to extensive offline resources for continual updates, an unrealistic expectation in resource-constrained environments. 
Parameter-Efficient Continual Learning (PECL) addresses this by leveraging Parameter-Efficient Fine-Tuning (PEFT), updating only a small subset of parameters. 
This significantly reduces overhead, making continual learning more practical in resource-constrained environments.
Despite its advantages, PECL still struggles with forgetting, as even minimal weight updates can interfere with previously learned tasks.
Although more efficient than full fine-tuning, PECL methods still involve time-consuming training per task, limiting their use in real-world scenarios. 
Given these challenges, the desideratum of continual learning is to eliminate forgetting while maintaining resource-efficient adaptation. 
While preventing weight updates entirely is the only way to fully avoid forgetting, it limits the model’s capacity to learn new tasks.
This trade-off presents a core challenge: \textit{How can we build a continual learning system that is both resource-efficient and capable of learning new tasks without forgetting?}
\newline\newline
In this work, we present Segment Anything with Incrementally Learned Semantics (SAILS), a novel training-free framework for Class-Incremental Semantic Segmentation (CISS). 
Our method leverages the powerful generalization capabilities of foundational models to segment objects and progressively learn new class semantics over time. 
We formulate CISS as a combination of two components: (1) Spatial segmentation and (2) Semantic association.
First, we utilize the Segment Anything Model (SAM) \cite{SAM} to extract object regions in a zero-shot, prompt-free manner.
Then, we incrementally assign semantic meaning to these segments using class prototypes derived from a frozen network pretrained on large and diverse data.
This enables continual adaptation without updating any model parameters.
Our approach offers several advantages:
\begin{itemize}
	\item \textbf{Training-Free Continual Learning: } Our framework enables continual learning without retraining, ensuring computational efficiency and near real-time adaptation.
	\item \textbf{Forgetting-Free Learning: }By avoiding model updates, SAILS sidesteps forgetting, maintaining stable performance across tasks without degradation over time. 
	\item \textbf{Task-Invariant Results: } SAILS achieves consistent performance regardless of task sequence or length.
	\item \textbf{Positive Backward Transfer: } SAILS promotes positive backward transfer, improving performance on previous classes while adapting to new classes.
	
\end{itemize}

\section{Background and Related Works}

Incremental learning involves learning from a sequence of tasks over time and is commonly categorized into three settings \cite{clScenarios}. 
In domain-incremental learning, the input distribution shifts while the number of classes remains fixed. 
In class-incremental learning, new classes from the same input distribution are introduced incrementally. 
Task-incremental learning can involve either domain or class shifts but assumes task IDs are available during inference. 
Recent works have extended these paradigms to more realistic scenarios: Evolving ontologies and semantic refinement \cite{LECO,CLEO}, incremental learning across modalities \cite{MIL,Harmony}, and simultaneous adaptation to domain and class shifts \cite{LWS,VIL}.
In this work, we focus on class-incremental semantic segmentation, where disjoint subsets of classes are introduced sequentially.

\begin{figure*}[t]
	\centering
	\includegraphics[width=\textwidth]{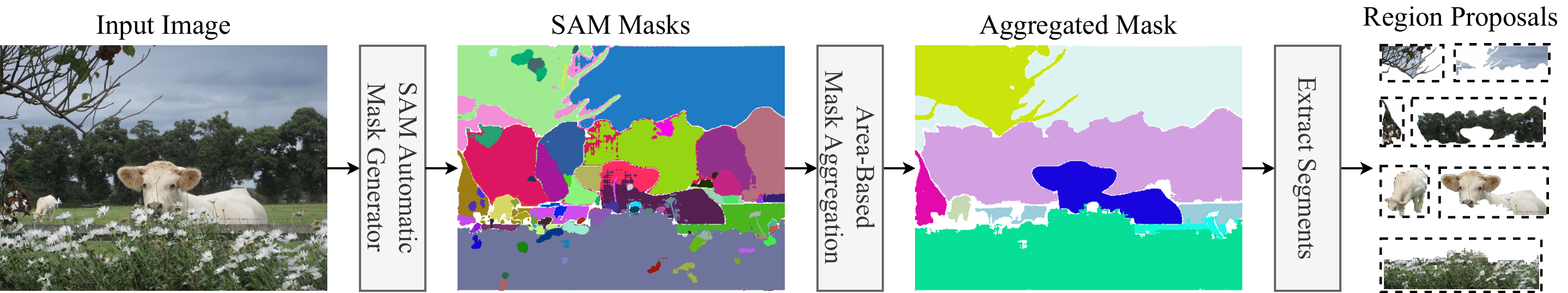}
	\caption{The input image is first segmented using SAM, and the resulting masks are iteratively refined to produce an aggregated mask with distinct, non-overlapping regions, which are then used for extraction of region proposals from the image.}
	\label{fig:mask_agg}
\end{figure*}

\subsection{Class-Incremental Semantic Segmentation}
ILT \cite{ILT} uses knowledge distillation across output and intermediate features with a frozen encoder. 
MiB \cite{MiB} mitigates background shift through a modified distillation loss, to compare previous background predictions with current background and new class predictions.
PLOP \cite{PLOP} uses pseudo-labels for background pixels representing previously seen classes. 
MBS \cite{MBS} highlights the misclassification of undetected old class pixels as background during pseudo-labeling, and proposes selective pseudo-labeling and adaptive distillation.
SSUL-M \cite{SSUL} introduces an unknown class to separate future classes from the background, in addition to pseudo-labeling of previous classes and rehearsal with exemplar memory.
ALIFE \cite{ALIFE} replays stored feature representations instead of explicitly storing data from previous tasks.
RECALL \cite{RECALL} retrieves or generates old class images and pseudo-labels them using previous models.
SATS \cite{SATS} leverages transformer self-attention maps and class-specific region pooling for both inter-class and intra-class knowledge distillation.
CIT \cite{CIT} introduces a class independent transformation to convert outputs into class-independent forms, enabling accumulative distillation.
NeST \cite {NEST} proposed a new classifier pre-tuning method that learns a transformation from old classifiers to initialize new ones. 

\subsection{Parameter-Efficient Continual Learning}

Parameter-Efficient Continual Learning (PECL) enables continual learning in resource-constrained environments by leveraging Parameter-Efficient Fine-Tuning (PEFT) methods to reduce trainable parameters and computational overhead.
PEFT methods \cite{peftSurvey} include additive tuning, which introduces new parameters through adapters \cite{adapterDrop, IA3}, or prompts \cite{promptTuning, prefixTuning}); Partial tuning of selected weights \cite{Bitfit, peftMasking, peftPruning}, and reparameterization with low-rank updates \cite{LoRA, DyLoRA}.
Building on these methods, recent PECL techniques include:
Training of task-specific LoRA models \cite{taskArithmeticLoRA} and merging of them using task arithmetic \cite{taskArithmetic}. 
CoLoR \cite{CoLoR} uses LoRA expert models and k-means clustering to infer task-ID during inference.
LAE \cite{LAE} introduces a three-stage framework: online PEFT for task learning, offline PEFT for knowledge accumulation, and ensembling for inference.
O-LoRA \cite{OLoRA} incrementally adds LoRA modules while enforcing orthogonality across tasks to reduce interference.
InfLoRA \cite{InfLoRA} proposes interference-free low-rank adaptation by constructing subspaces to prevent interference between tasks.
CLoRA \cite{CLoRA} uses a single LoRA module with knowledge distillation to transfer knowledge between tasks in class-incremental semantic segmentation. \looseness-1

\subsection{Continual Learning with Pre-trained Networks}
Recent advancements in foundation models and large-scale pretraining have significantly influenced continual learning.
Pretrained models provide strong generalization capabilities, reducing the need for extensive task-specific training.
However, adapting these models for continual learning introduces unique challenges, including catastrophic forgetting of pretrained knowledge, distribution shifts between pretraining and downstream tasks, and limited plasticity in frozen or partially frozen networks.
Pelosin \cite{cl_pretrained_classification} presents a training-free approach that leverages pretrained models to compute class prototypes and populate a memory bank. 
Similarly, \cite{simple_pretrained} demonstrate that a frozen pretrained feature extractor with a Nearest Mean Classifier (NMC) can outperform more complex methods, underscoring the strength of pretrained representations.
Extending this idea, \cite{APER} proposes Adapt and Merge (APER), which first adapts the pretrained model on the initial training set using PEFT to bridge the domain gap. 
APER concatenates frozen embeddings from both models to preserve generalizability and adaptivity.
Supervised Contrastive Replay \cite{SCR} improves nearest-prototype classification by encouraging intra-class compactness and inter-class separation in the embedding space.
RanPAC \cite{RanPAC} inserts a frozen random projection layer between pretrained model features and the output head to improve separability between classes, ensuring better class prototypes \looseness-1

\section{Segment Anything with Incrementally Learned Semantics}

While continual learning with pretrained models has made significant strides, the majority of these advancements have focused primarily on image classification tasks. 
Class-incremental learning involves incrementally learning from a sequence of tasks $T = \{0, 1, \dots, n\} $, where each task $t$ introduces non-overlapping subsets of classes such that $C_t \subset C$ and $C_i \cap C_j = \emptyset$ $\forall  i \neq j$.
In classification, each image $I$ corresponds to a single label $c \in C_t$, and the subset of classes $C_t \subset C$ encountered across tasks $T$ is disjoint.
In contrast, semantic segmentation involves dense, pixel-wise predictions where multiple classes may exist within a single image.
At each time step $t$, the model has access to annotations only belonging to a subset of classes $C_t$. Pixels belonging to classes outside $C_t$, which include classes seen so far $C_{0:t-1}$ and potential future classes $C_{t+1:n}$ are labeled as background.
When distilling knowledge between tasks, this results in a mismatch between the predictions of the previous and current task model, resulting in background shift and exacerbating the problem of catastrophic forgetting.
Background shift necessitates relearning of the current classes, which may have been previously learned as the background class, and would require retraining. 
In a training-free setup, where updating previous knowledge is not possible, we must find a way to handle classes outside the current subset $C_t$ without retraining.
To address this challenge, we propose a modular decomposition of class-incremental semantic segmentation into two subproblems:

\begin{itemize}
	
	\item \textbf{Spatial Segmentation}
	Using SAM, we generate class-agnostic region proposals $\mathcal{R} = \{r_1, r_2, \dots, r_K\}$. These proposals represent potential objects and provide a spatial prior for downstream semantic reasoning.
	
	\item \textbf{Semantic Association}
	As new classes $C_t$ are introduced, we update the semantic understanding of the regions using class prototypes, ensuring a consistent and continually refined mapping from regions to classes.

\end{itemize}

\begin{figure*}[t]
	\centering
	\includegraphics[width=0.8\textwidth]{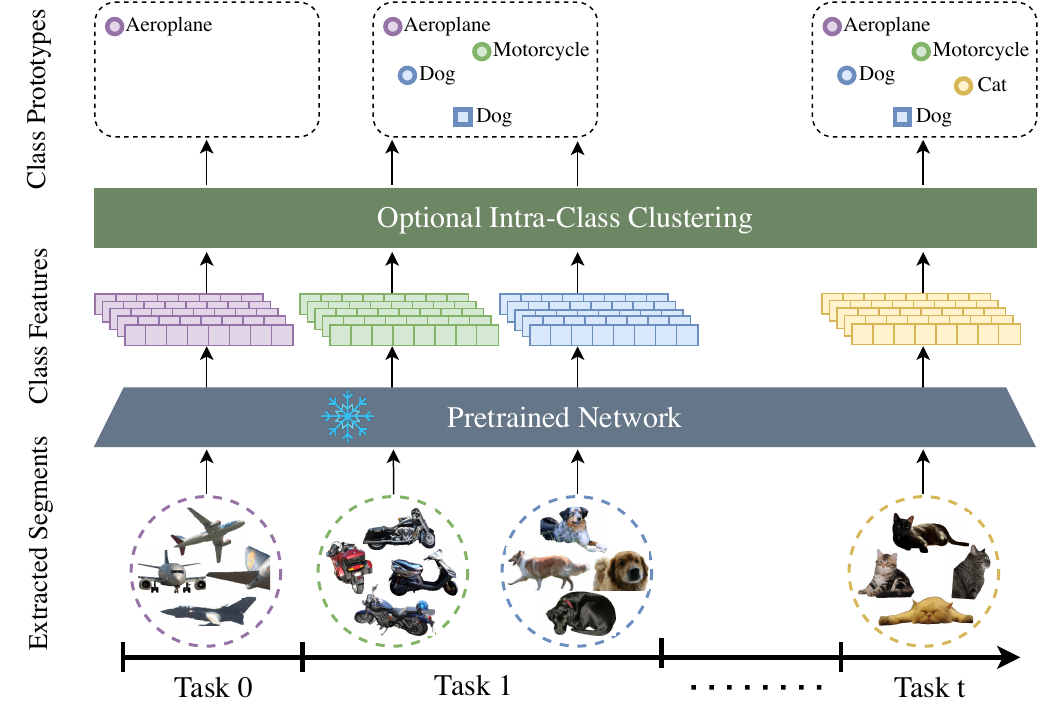} 
	\caption{Overview of the incremental semantic learning process. Regions of interest (RoIs) corresponding to classes introduced in the current task are embedded using a frozen pretrained backbone. For each class, either a single prototype or multiple sub-prototypes are computed based on the intra-class variability.}
	\label{fig:semantics}
\end{figure*}

\subsection{Spatial Segmentation}

We first segment the input images using the automatic mask generator from Segment Anything Model (SAM) \cite{SAM}.
SAM provides the flexibility to generate segmentation masks without any task-specific training, making it suitable for continual learning. 
However, integrating SAM into class-incremental semantic segmentation poses two key challenges: mask granularity and missing semantic context.

\subsubsection{Mask Granularity.} 
The output from SAM typically consists of a large number of masks, many of which are overlapping and redundant. 
These masks tend to be fine-grained, often representing sub-parts or fragments of objects.
While this granularity can be valuable for certain applications, it poses a challenge for semantic segmentation. 
These fragmented regions do not adequately capture the holistic characteristics of objects, which is crucial for subsequent semantic learning.
To address these challenges, we implement a refinement step that consolidates the initial SAM output into a single mask per image. 
By prioritizing large, non-overlapping masks and discarding redundant and partial masks, our approach produces object-level segments.
\newline\newline
Let \( I \) be the input image, and  \( \mathcal{M} = \{ M_1, M_2, \ldots, M_N \} \) denote the set of binary masks generated by SAM, where each mask \( M_i \in \{0,1\}^{H \times W} \) represents a region and is accompanied by metadata, including its area.
We first remove masks that trivially cover most of the image, as they are unlikely to represent meaningful object instances. 
We retain masks whose area is less than a threshold fraction $\tau_{area} = 0.9$ of the total image size: 
\begin{equation}
\mathcal{M}' = \{ M_i \in \mathcal{M} \mid \text{area}_i < \tau_{area} \cdot (H \cdot W) \}
\end{equation}
\newline
Next, we sort the filtered masks in descending order by area. Larger masks are more likely to represent entire objects, whereas smaller masks often correspond to object parts.
Sorting ensures that larger, more comprehensive regions are prioritized during mask consolidation:
\begin{equation}
\mathcal{M}' = \{ M_1, M_2, \ldots, M_K \} \hspace{0.5em} \text{with}  \hspace{0.5em}  A_1 \geq \cdots \geq A_K
\end{equation}
where \( A_i \) denotes the area of mask \( M_i \).
\newline\newline
We initialize an empty composite mask \( M_{\text{agg}} \in \{1, 2, \ldots \}^{H \times W} \), where each pixel holds an integer label corresponding to a region \(r_k\). Initially, all values in \( M_{\text{agg}} \) are zero, indicating that no region has been assigned.
We then iterate over each sorted mask \( M_i \in \mathcal{M}' \). For each mask, we check if it contains foreground pixels not yet labeled in \( M_{\text{agg}} \). If such pixels exist, we assign a new label \( c \) to these pixels, incrementing \( c \) sequentially for each new region.
The update rule for each pixel \( (x, y) \) is:
\begin{equation} 
	M_{\text{agg}}(x, y) = c \quad \text{if } M_i(x, y) = 1 \text{ and } M_{\text{agg}}(x, y) = 0
\end{equation}
We repeat this process for all masks in \(\mathcal{M}'\), resulting in the composite mask \( M_{\text{agg}} \) that contains unique and non-overlapping region proposals \(\mathcal{R} = \{ r_1, r_2, \ldots, r_K \}\).  
These regions form the basis for the subsequent incremental semantic learning.
\cref{fig:mask_agg} illustrates this aggregation process, where the output from SAM is consolidated into \( M_{\text{agg}} \).

\subsection{Semantic Association}

SAM produces class-agnostic masks by segmenting image regions purely based on visual and spatial characteristics, without any semantic information. 
Following the initial segmentation step using SAM, the resulting mask segments or region proposals are used for extracting regions of interest (RoIs) from the image. 
The key challenge that follows is to incrementally assign semantic labels to these segments as new classes are introduced over time.
This requires a continual learning method that sequentially integrates new classes with unlabeled regions while preserving previously learned knowledge and preventing forgetting.
This reframes continual semantic segmentation as a continual classification problem of the segmented regions. 
Although existing continual learning methods for image classification could be adapted for this purpose, they rely on training and updating the model, which brings us back to the core motivation of this paper: Training is computationally expensive, forgetting becomes inevitable, and severe performance degradation often occurs over long task sequences.

To overcome these challenges, we propose a prototype-based classifier built upon frozen, pretrained vision backbones that requires no additional training or parameter updates. Our approach consists of the following steps:

\begin{itemize}
	\item \textbf{Feature Extraction:}  Given an input image, we extract Regions of Interest (RoIs) using the aggregated SAM mask. Each RoI \( r \) is a rectangular crop around the object, with the background masked and is passed through a frozen network \( f(\cdot) \) to obtain a feature embedding:
	\begin{equation}
	   \mathbf{z}_r = norm(f(r)) \in \mathbb{R}^d    
	\end{equation}	
	where \( d \) is the feature dimension.
	
	\item \textbf{Prototype Computation:} For each new class \( c \in C_t \), we compute a class prototype \(\mathbf{p}_c\) by averaging the embeddings of all segments belonging to class \( c \):
	\begin{equation}
	   \mathbf{p}_c = norm(\frac{1}{N_c} \sum_{i=1}^{N_c} \mathbf{z}_{r_i}),
	\end{equation}
	where \( N_c \) is the number of segments labeled as class \( c \).
	
	\item \textbf{Classification by Prototype Similarity:} To classify a new segment \( r \), we compute the cosine similarity between its embedding $z_r$ and each stored prototype for all classes  \( \mathcal{C}_{0:t} \) encountered up to task \( t \). The segment is assigned the class with the highest similarity score exceeding a predefined threshold \( \tau_{sim} \):
    \begin{equation}
	y_r = 
	\begin{cases}
		\displaystyle \arg\max_{c \in \mathcal{C}_{0:t}} \frac{\mathbf{z}_r \cdot \mathbf{p}_c}{\|\mathbf{z}_r\|\|\mathbf{p}_c\|}, & \text{if } sim \geq \tau_{sim} \\
		\text{BG}, & \text{otherwise}
	\end{cases}
    \end{equation}
\end{itemize}

To address the background shift problem in class-incremental semantic segmentation, we incorporate a similarity threshold \( \tau_{sim} = 0.5 \) during classification. 
Some RoIs may correspond to future classes or background regions that are not semantically relevant.
As both cases lack valid labels in the current task, we assign such RoIs to the background (BG) class if their similarity to all known class prototypes falls below the threshold.
This prevents misclassification and avoids assigning incorrect semantics to ambiguous or currently irrelevant regions.
\subsubsection{Intra-Class Variance}
A key limitation of relying on a single prototype per class is its inability to capture high intra-class variability effectively.
Classes like dogs with varied breeds or vehicles of different types exhibit significant visual diversity that a single prototype cannot fully represent.
Averaging widely dispersed embeddings can cause the class prototype to fall into semantically ambiguous regions of the feature space, failing to align well with any specific subgroup. 
Addressing this limitation, we introduce an intra-class clustering mechanism for classes with high variance.
To identify classes with high variance, we compute a variance score based on cosine distance.
For each class, we calculate the mean feature vector $\mu_c$ and measure the average cosine distance to all feature vectors in the class. 
This score reflects how tightly the features are grouped, with higher values indicating more variance.
Classes with variance scores above 0.4 are chosen for sub-clustering.
\begin{equation}
\text{Variance}_c = \frac{1}{N_c} \sum_{i=1}^{N_c} \text{cosine\_distance}(\mathbf{x}_i, \mu_c)
\end{equation}
By freezing the backbone and only incrementally adding new class prototypes, our approach remains training-free, efficient, and inherently resistant to forgetting as it avoids interference with previous class prototypes.
\cref{fig:semantics} illustrates this process: RoIs corresponding to classes introduced in the current task are embedded using a frozen pretrained backbone, and prototypes are computed for each class.
During inference, the RoI embedding is matched against all prototypes and sub-prototypes across classes and assigned the class with the highest similarity. \looseness-1

\section{Results}
In this section, we present experiments and results that demonstrate the effectiveness of our approach. 
We begin by describing the datasets used, followed by the baseline methods and implementation details. 
We then provide a detailed analysis of our approach’s performance across incremental tasks, highlighting its advantages over existing methods.

\subsection{Datasets}
We evaluate our approach on two standard benchmarks for class-incremental semantic segmentation (CISS):

\begin{itemize}
	\item \textbf{PASCAL VOC 2012 \cite{PASCAL}} consists of 21 semantic classes with 1,464 images in the training set. 
    Most CISS methods use the augmented set with over 10k images, while our training-free approach achieves similar results using only the original training split, eliminating the need for large-scale annotated data.
	
	\item \textbf{Cityscapes \cite{cityscapes}} is a challenging urban scene understanding dataset with 19 semantic classes and 2,975 training images.
	Our results on Cityscapes further highlight the effectiveness and suitability of our approach for real-world applications like autonomous driving.
\end{itemize}

\subsection{Baselines and Implementation}
For all comparisons, we adopt the SegFormer-B2 architecture from the SATS framework \cite{SATS}, while for SSUL, we use its corresponding framework.
We perform a comprehensive comparison against joint training (JT) and several state-of-the-art CISS methods, including ILT \cite{ILT}, MiB \cite{MiB}, PLOP \cite{PLOP}, SSUL \cite{SSUL}, and SATS \cite{SATS}. 
The joint training baseline trains the model on all classes simultaneously in a single task. 
As a result, it does not experience catastrophic forgetting and serves as the upper-bound for comparison.
We also compare against SAM+CLIP, which combines SAM and the default CLIP \cite{CLIP} ViT-B model with the officially published code for zero-shot prediction.
While it shares the training-free benefits, we highlight its challenges and limitations.
For feature extraction in our approach, we use a Swin-B \cite{Swin} network pretrained on ImageNet. 
To handle classes with high variance, we apply k-means clustering with \(k=5\) within each class to obtain multiple fine-grained sub-prototypes that better capture semantic diversity. \looseness-1 

\begin{table}[t]
    \centering
    \caption{Results (in mIoU) on PASCAL VOC \cite{PASCAL} dataset after learning all steps. SAILS$^{*}$ denotes intra-class clustering.}
	\label{tab:voc}
	\begin{adjustbox}{width=\columnwidth}
		\centering
		\begin{tabular}{c||c|c|c|c}
			\boldhline
			\multirow{2}{*}{\textbf{Method}} & \textbf{1-1} & \textbf{2-1} & \textbf{5-1} & \textbf{2-2} \\ 
			& (20 Steps) & (19 Steps) & (16 Steps) & (10 Steps) \\ \boldhline
			JT  			 & 80.12 & 80.12 & 80.12 & 80.12 \\ \boldhline
			ILT \cite{ILT}   & 05.55 & 05.95 & 06.25 & 10.72 \\ \hline
			MiB \cite{MiB}   & 17.03 & 25.74 & 45.63 & 51.83 \\ \hline
			PLOP \cite{PLOP} & 07.55 & 08.14 & 06.94 & 11.32 \\ \hline
			SATS \cite{SATS} & 17.97 & 05.89 & 37.86 & 39.74 \\ \hline
			SSUL \cite{SSUL} & 29.15 & 38.32 & 48.65 & 45.31 \\ \boldhline
			CLIP             & 35.97 & 35.97 & 35.97 & 35.97  \\ \hline
			SAILS\phantom{$^{*}$}   & 54.43 & 54.43 & 54.43 & 54.43 \\ \hline
			SAILS$^{*}$      & \textbf{57.96} & \textbf{57.96} & \textbf{57.96} & \textbf{57.96} \\ \boldhline
		\end{tabular}
	\end{adjustbox}
	
\end{table}

\subsection{PASCAL VOC}
We evaluate robustness under long task sequences where conventional methods typically struggle due to the cumulative effect of forgetting.
CIL settings use \textit{init-inc} notation, where \textit{init} is the initial number of classes and \textit{inc} the classes added incrementally, both of which influence the severity of forgetting.
We conduct experiments under the \textit{1-1, 2-1, 5-1} and \textit{2-2} task splits, which lead to a large number of incremental steps.
The performance of our approach is compared against state-of-the-art (SotA) baselines, as summarized in \cref{tab:voc}.
Performance of existing approaches degrades significantly as the number of steps increases, highlighting their limitations in handling long task sequences. 
In contrast, our training-free framework, SAILS, maintains consistent performance across all task configurations, showcasing its robustness and task-invariant results regardless of sequence length.
Furthermore, with intra-class clustering, we observe improved performance across tasks. \looseness-1

\begin{table}[t]
    \centering
    \caption{Results (in mIoU) on Cityscapes \cite{cityscapes} dataset after learning all tasks. SAILS$^{*}$ denotes intra-class clustering.}
	\label{tab:cityscapes}
	\begin{adjustbox}{width=\columnwidth}
		\centering
		\begin{tabular}{c||c|c|c|c}
			\boldhline
			\multirow{2}{*}{\textbf{Method}} & \textbf{1-1} & \textbf{2-1} & \textbf{5-1} & \textbf{3-2} \\ 
			& (19 Steps) & (18 Steps) & (15 Steps) & (9 Steps) \\ \boldhline
			JT  			 & 59.06 & 59.06 & 59.06 & 59.06 \\ \boldhline
			ILT  \cite{ILT}  & 00.12 & 00.22 & 01.52 & 00.81 \\ \hline
			MiB \cite{MiB}   & 29.71 & 31.03 & 35.35 & 35.06 \\ \hline
			PLOP \cite{PLOP} & 11.48 & 11.78 & 16.79 & 22.10 \\ \hline
			SATS \cite{SATS} & 15.16 & 21.26 & 24.12 & 33.63 \\ \boldhline
                CLIP             & 15.48 & 15.48 & 15.48 & 15.48  \\ \hline
			SAILS\phantom{$^{*}$}  & 36.51 & 36.51 & 36.51 & 36.51 \\ \hline
			SAILS$^{*}$      & \textbf{45.06} & \textbf{45.06} & \textbf{45.06} & \textbf{45.06}\\ \boldhline
		\end{tabular}
	\end{adjustbox}
\end{table}

\begin{figure*}[ht]
	\centering
	\begin{subfigure}[b]{0.24\textwidth}
		\centering
		\includegraphics[width=\textwidth]{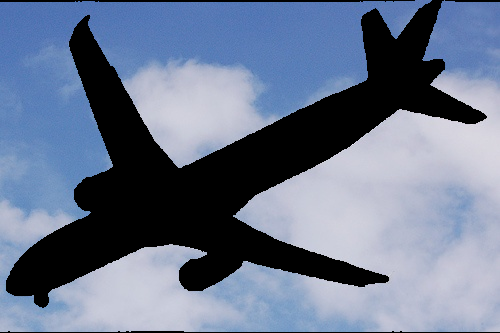}
		\caption*{Predicted: \textit{Aeroplane}}
	\end{subfigure}
	\hfill
	\begin{subfigure}[b]{0.24\textwidth}
		\centering
		\includegraphics[width=\textwidth]{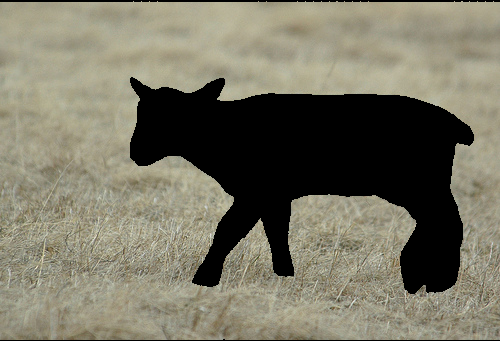}
		\caption*{Predicted: \textit{Cow}}
	\end{subfigure}
	\hfill
	\begin{subfigure}[b]{0.24\textwidth}
		\centering
		\includegraphics[width=\textwidth]{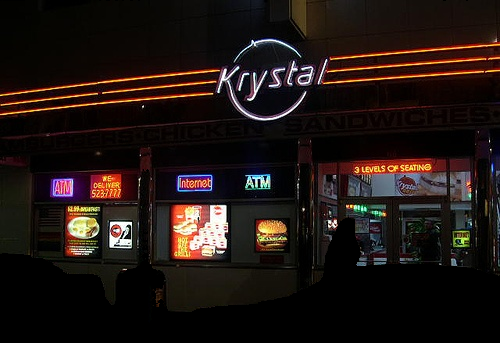}
		\caption*{Predicted: \textit{Monitor}}
	\end{subfigure}
		\hfill
	\begin{subfigure}[b]{0.24\textwidth}
		\centering
		\includegraphics[width=\textwidth]{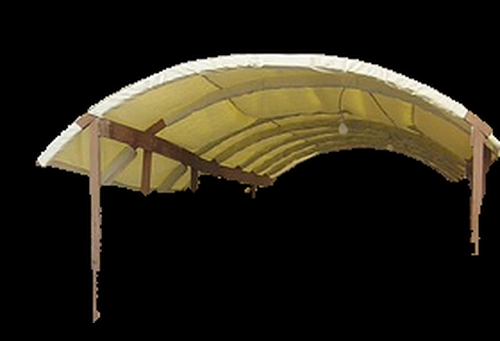}
		\caption*{Predicted: \textit{Diningtable}}
	\end{subfigure}
	\caption{CLIP misclassifications of region segments due to shortcut learning and ambiguous regions. }
	\label{fig:clip-misclassifications}
\end{figure*}

\begin{table}[t]
	\centering
	\caption{Comparison of segment quality across methods.}
	\label{tab:segmentation_quality}
        \begin{tabular}{c||>{\centering\arraybackslash}p{1.2cm}|>{\centering\arraybackslash}p{1.2cm}|>{\centering\arraybackslash}p{1.2cm}}
		\boldhline
		\textbf{Segments} & \textbf{mIoU} & \textbf{Precision} & \textbf{Recall} \\ \boldhline
		Ground Truth     & 56.18  & 80.35    & 71.20    \\ \hline 
        Ground Truth (Non-BG)  & 70.87  & 98.49    & 71.64    \\ \hline
        SAM  \cite{SAM} & 27.52  & 47.67    & 37.67    \\ \hline
     	Mask2Former \cite{Mask2Former}  & 40.51  & 64.71    & 48.14    \\ \hline
		SAM-Agg (Ours) & 54.43  & 79.25    & 63.67    \\ \boldhline
	\end{tabular}
\end{table}

\begin{table*}[t]
    \centering
    \caption{Task-invariant class-wise IoU on PASCAL VOC \cite{PASCAL}. SAILS$^{*}$ denotes our approach with intra-class clustering.
    Classes represented with intra-class clustering are highlighted in \textbf{bold}.}
	\label{tab:clustering}
	\begin{adjustbox}{width=\textwidth}
		\centering
		\begin{tabular}{c||*{21}{c}}
			\boldhline
			\textbf{Method} & 
			\rotatebox{60}{background} & 
			\rotatebox{60}{aeroplane} & 
			\rotatebox{60}{bicycle} & 
			\rotatebox{60}{\textbf{bird}} & 
			\rotatebox{60}{\textbf{boat}} & 
			\rotatebox{60}{\textbf{bottle}} & 
			\rotatebox{60}{bus} & 
			\rotatebox{60}{\textbf{car}} & 
			\rotatebox{60}{cat} & 
			\rotatebox{60}{\textbf{chair}} & 
			\rotatebox{60}{cow} & 
			\rotatebox{60}{\textbf{diningtable}} & 
			\rotatebox{60}{\textbf{dog}} & 
			\rotatebox{60}{horse} & 
			\rotatebox{60}{motorbike} & 
			\rotatebox{60}{\textbf{person}} & 
			\rotatebox{60}{\textbf{pottedplant}} & 
			\rotatebox{60}{sheep} & 
			\rotatebox{60}{\textbf{sofa}} & 
			\rotatebox{60}{train} & 
			\rotatebox{60}{tvmonitor} \\
			\boldhline
			SAILS\phantom{$^{*}$}         & 83.13 & 32.13 & 42.63 & 34.53 & 52.88 & 52.40 & 88.16 & 60.24 & 83.59 & 34.65 & 77.49 & 01.86 & 17.48 & 87.28 & 78.60 & 32.76 & 34.64 & 70.81 & 25.94 & 81.68 & 70.17 \\
			SAILS$^{*}$   & 83.16 & 32.13 & 42.63 & \cellcolor{green!20} 41.20 & \cellcolor{green!20} 64.84 & \cellcolor{green!20} 58.37 & 88.16 & \cellcolor{green!20} 73.09 & 83.59 & \cellcolor{red!20} 24.12 & 77.46 & \cellcolor{green!20} 10.51 & \cellcolor{green!20} 42.63 & 87.28 & 78.82 & \cellcolor{green!20} 55.12 & \cellcolor{red!20} 25.78 & 70.81 & \cellcolor{red!20} 25.65 & 81.68 & 70.17 \\
			\boldhline
		\end{tabular}
	\end{adjustbox}
\end{table*}

\begin{table*}[t]
\centering
    \caption{Class-wise IoU without intra-class clustering on PASCAL VOC \cite{PASCAL} (15-5 setting), highlighting positive backward transfer on the initial 15 classes after learning the final 5 classes in step 1.}
	\label{tab:backward-transfer}
	\begin{adjustbox}{width=\textwidth}
		\centering
		\begin{tabular}{c||*{21}{c}}
			\boldhline
			\textbf{Step} & 
			\rotatebox{60}{background} & 
			\rotatebox{60}{aeroplane} & 
			\rotatebox{60}{bicycle} & 
			\rotatebox{60}{bird} & 
			\rotatebox{60}{boat} & 
			\rotatebox{60}{bottle} & 
			\rotatebox{60}{bus} & 
			\rotatebox{60}{car} & 
			\rotatebox{60}{cat} & 
			\rotatebox{60}{chair} & 
			\rotatebox{60}{cow} & 
			\rotatebox{60}{diningtable} & 
			\rotatebox{60}{dog} & 
			\rotatebox{60}{horse} & 
			\rotatebox{60}{motorbike} & 
			\rotatebox{60}{person} & 
			\rotatebox{60}{pottedplant} & 
			\rotatebox{60}{sheep} & 
			\rotatebox{60}{sofa} & 
			\rotatebox{60}{train} & 
			\rotatebox{60}{tvmonitor} \\
			\boldhline
			0 & 
			86.16 & 32.13 & 42.63 & 34.53 & 52.88 & 52.40 & 86.47 & 60.24 & 83.59 & 33.06 & 77.11 & 01.89 & 17.47 & 87.28 & 78.60 & 32.75 &
			\cellcolor{gray!60} & 
			\cellcolor{gray!60} & 
			\cellcolor{gray!60} & 
			\cellcolor{gray!60} & 
			\cellcolor{gray!60}  
			\\
			
			1 & 
			 83.13 & 32.13 & 42.63 & 34.53 & 52.88 & 52.40 & \textbf{88.16} & 60.24 & 83.59 & \textbf{34.65} & \textbf{77.49} & 01.86 & 17.48 & 87.28 & 78.60 & 32.76 & 34.64 & 70.81 & 25.94 & 81.68 & 70.17 \\ 

			\boldhline
		\end{tabular}
	\end{adjustbox}
\end{table*}

\subsection{Cityscapes}
To further validate the generalizability of our approach, we evaluate SAILS on the Cityscapes \cite{cityscapes} dataset under similar challenging task splits: \textit{1-1, 2-1, 3-2}, and \textit{5-1}.
We compare SAILS against SotA baselines under these extreme task splits and the results are presented in \cref{tab:cityscapes}.
Forgetting is amplified by the underrepresented classes such as \textit{motorcycle} and \textit{bicycle}, which are encountered toward the end of the task sequence in Cityscapes, and hinders effective learning in later stages.
In contrast, SAILS exhibits stable performance across all tasks, mirroring the robustness observed on PASCAL VOC.
Our training-free approach circumvents the cumulative performance degradation observed in methods that rely on model updates. \looseness-1

\subsection{Intra-Class Clustering}
While our training-free approach ensures efficiency and avoids forgetting, it relies on frozen features, limiting adaptability to intra-class variability.
To address this, we introduce \textit{intra-class clustering}, allowing multiple prototypes per class to better capture feature diversity. 
As shown in \cref{tab:clustering}, intra-class clustering leads to improved segmentation performance for most of these classes.
However, for visually similar classes like \textit{chair} and \textit{sofa}, performance slightly drops. 
This is likely due to increased confusion between these classes, as the additional prototypes can represent similar visual features.
For \textit{potted plant}, clustering improves recall but reduces precision. \textit{Background} segments like trees and vegetation are misclassified as potted plants due to their visual similarity. \looseness-1

\subsection{Positive Backward Transfer}

While most continual learning methods have primarily focused on mitigating forgetting, SAILS goes further by exhibiting positive backward transfer, improving performance on previous tasks as new classes are added.
This effect is more pronounced when the newly added classes are visually similar to the previously learned classes.
The new semantic context helps disambiguate earlier misclassifications between visually similar classes. 
We illustrate this phenomenon in \cref{tab:backward-transfer} using the \textit{15-5} task setting on PASCAL VOC. 
After learning the initial 15 classes in step 0, we observe that their performance improves following the introduction of the final 5 classes in step 1. 
For instance, introducing \textit{train} enhances the previously learned \textit{bus} by offering finer semantic contrast. 
Similarly, the addition of \textit{sofa} improves the visually similar \textit{chair} class. \looseness-1

\subsection{Zero-Shot Segmentation using SAM and CLIP}
While CLIP \cite{CLIP} is effective for zero-shot classification, applying it to region-level segmentation reveals several limitations.
It exhibits shortcut learning, where regions are misclassified based on scene context rather than object-specific features.
In \cref{fig:clip-misclassifications}, the sky surrounding the plane is misclassified as \textit{aeroplane}, likely due to the plane's silhouette.
Similarly, the grass around the \textit{sheep} is labeled as \textit{cow}, likely due to the contextual similarity between the two scenes.
CLIP also struggles with partial or ambiguous region segments such as windows and signboards being classified as \textit{monitor} based on shape similarity.
More critically, CLIP struggles in continual learning settings that require recognizing novel, fine-grained classes not encountered during its pretraining. 
Without task-specific fine-tuning, its utility for incremental learning is limited. \looseness-1

\section{Discussion}
In this section, we discuss key design choices and limitations of our framework, and outline promising directions for future work. 
Our goal is to highlight the trade-offs underlying our approach and identify areas where further research could extend its applicability. \looseness-1  

\subsection{Leveraging SAM for Segmentation}

A key design choice in our framework is the use of SAM \cite{SAM} for universal region segmentation. 
While computationally heavier, SAM provides dense and high-quality masks and crucially, generalizes across domains without retraining.
In contrast, salient object detection covers only dominant foreground objects, and panoptic segmentation models are dataset and domain specific, leading to degraded performance when applied to new domains.
To quantify segmentation quality, we compare SAM segments, our aggregated object-level segments (SAM-Agg), and ground truth (GT). 
We include proposals from \cite{MicroSeg} generated using Mask2Former \cite{Mask2Former}, a panoptic segmentation model trained on MS-COCO \cite{COCO}.
We assign classes using the same training-free, prototype-based classification, and the results are presented in \cref{tab:segmentation_quality}.
\newline\newline
Ground-truth segments represent the upper bound of segmentation performance, providing precise and semantically consistent masks, serving both as a reference for ideal segmentation and evaluating our prototype-based classification.
Excluding background segments further improves accuracy and precision while maintaining high recall.
SAM’s fine-grained masks capture detailed part-level structures but often over-segment objects into fragmented components such as parts or limbs. 
Our refinement (SAM-Agg) consolidates these into coherent object-level masks, leading to an improvement in precision and recall while also reducing the number of segments and thus computational cost. 
Mask2Former produces stronger segments than SAM, but this advantage is largely due to its training on COCO \cite{COCO}, which closely resembles PASCAL VOC \cite{PASCAL} highlighting SAM’s strength as a training-free, domain-agnostic alternative. 
More computationally efficient alternatives to SAM, can help alleviate the computational demands and lower resource requirements.

\subsection{Frozen Pretrained Representations}

Since our framework uses a frozen feature extractor, its adaptability is ultimately constrained by the discriminative capacity of the pretrained backbone.
To mitigate this limitation, we employ selective intra-class clustering to identify subgroups within each class. 
However, distinguishing between visually or semantically similar classes remains challenging. 
Classes such as \textit{sofa} and \textit{chair} share many visual characteristics, leading to frequent misclassifications.
Using a frozen backbone without task-specific retraining ensures task-invariance and prevents catastrophic forgetting, but it also constrains adaptability.
Future work could explore methods to improve downstream adaptation while retaining efficiency through PECL. 
Additionally, methods built on top of the frozen features, such as contrastive learning, can enhance separability between classes while maintaining the benefits of a frozen, stable backbone.\looseness-1 

\section{Conclusion}
Continual learning is often hindered in real-world settings due to the need for repeated retraining, high computational costs, and catastrophic forgetting.
Addressing these challenges, we present SAILS, a training-free framework for class-incremental semantic segmentation that avoids retraining and forgetting by leveraging SAM for zero-shot region extraction and prototype-based semantic association. 
With selective intra-class clustering, SAILS effectively handles class variability and consistently outperforms training-based methods, especially in long task sequences. 
The training-free approach ensures task-invariant performance and even positive backward transfer, making it a robust and efficient solution for continual semantic segmentation.

\section*{Acknowledgments}
This work was partially funded by the German Federal Ministry of Research, Technology, and Space under the project COPPER (16IW24009).

{
    \small
    \bibliographystyle{ieeetr}
    \bibliography{main}

@article{isolated_learning,
	title={Lifelong machine learning: a paradigm for continuous learning},
	author={Liu, Bing},
	journal={Frontiers of Computer Science},
	year={2017}}

@article{catastrophic_forgetting,
	title={Catastrophic forgetting in connectionist networks},
	author={French, Robert M},
	journal={Trends in cognitive sciences},
	year={1999}}

@misc{stability_plasticity_dilemma,
  title={The stability-plasticity dilemma: Investigating the continuum from catastrophic forgetting to age-limited learning effects},
  author={Mermillod, Martial and Bugaiska, Aur{\'e}lia and Bonin, Patrick},
  journal={Frontiers in Psychology},
  year={2013}}

@article{clScenarios,
	title={Three scenarios for continual learning},
	author={Van de Ven, Gido M and Tolias, Andreas S},
	journal={arXiv preprint arXiv:1904.07734},
	year={2019}}

@inproceedings{CLEO,
	title={CLEO: Continual Learning of Evolving Ontologies},
	author={Muralidhara, Shishir and Bukhari, Saqib and Schneider, Georg and Stricker, Didier and Schuster, Ren{\'e}},
	booktitle={European Conference on Computer Vision (ECCV)},
	year={2024}}

@article{LECO,
  title={Continual learning with evolving class ontologies},
  author={Lin, Zhiqiu and Pathak, Deepak and Wang, Yu-Xiong and Ramanan, Deva and Kong, Shu},
  journal={Advances in Neural Information Processing Systems (NeurIPS)},
  year={2022}}

@inproceedings{MIL,
	title={Modality-Incremental Learning with Disjoint Relevance Mapping Networks for Image-based Semantic Segmentation},
	author={Hegde, Niharika and Muralidhara, Shishir and Schuster, Ren{\'e} and Stricker, Didier},
	booktitle={Winter Conference on Applications of Computer Vision (WACV)},
	year={2025}}

@article{Harmony,
  title={Harmony: A Unified Framework for Modality Incremental Learning},
  author={Song, Yaguang and Yang, Xiaoshan and Jiang, Dongmei and Wang, Yaowei and Xu, Changsheng},
  journal={arXiv:2504.13218},
  year={2025}
}

@article{LWS,
	title={Learning with style: Continual semantic segmentation across tasks and domains},
	author={Toldo, Marco and Michieli, Umberto and Zanuttigh, Pietro},
	journal={IEEE Transactions on Pattern Analysis and Machine Intelligence},
	year={2024}}

@inproceedings{VIL,
  title={Versatile Incremental Learning: Towards Class and Domain-Agnostic Incremental Learning},
  author={Park, Min-Yeong and Lee, Jae-Ho and Park, Gyeong-Moon},
  booktitle={European Conference on Computer Vision},
  pages={271--288},
  year={2024},
  organization={Springer}
}

@inproceedings{ILT,
	title={Incremental learning techniques for semantic segmentation},
	author={Michieli, Umberto and Zanuttigh, Pietro},
	booktitle={International Conference on Computer Vision Workshops (ICCVW)},
	year={2019}}

@inproceedings{MiB,
	title={Modeling the background for incremental learning in semantic segmentation},
	author={Cermelli, Fabio and Mancini, Massimiliano and Bulo, Samuel Rota and Ricci, Elisa and Caputo, Barbara},
	booktitle={Conference on Computer Vision and Pattern Recognition (CVPR)},
	year={2020}
}

@inproceedings{PLOP,
	title={Plop: Learning without forgetting for continual semantic segmentation},
	author={Douillard, Arthur and Chen, Yifu and Dapogny, Arnaud and Cord, Matthieu},
	booktitle={Conference on Computer Vision and Pattern Recognition (CVPR)},
	year={2021}}

@article{SATS,
	title={SATS: Self-attention transfer for continual semantic segmentation},
	author={Qiu, Yiqiao and Shen, Yixing and Sun, Zhuohao and Zheng, Yanchong and Chang, Xiaobin and Zheng, Weishi and Wang, Ruixuan},
	journal={Pattern Recognition},
	year={2023}}

@article{SSUL,
	title={SSUL: Semantic segmentation with unknown label for exemplar-based class-incremental learning},
	author={Cha, Sungmin and Yoo, YoungJoon and Moon, Taesup and others},
	journal={Advances in Neural Information Processing Systems (NeurIPS)},
	year={2021}}

@inproceedings{ALIFE,
	author = {Oh, Youngmin and Baek, Donghyeon and Ham, Bumsub},
	booktitle = {Advances in Neural Information Processing Systems (NeurIPS)},
	title = {ALIFE: Adaptive Logit Regularizer and Feature Replay for Incremental Semantic Segmentation},
	year = {2022}}

@inproceedings{RECALL,
	title={Recall: Replay-based continual learning in semantic segmentation},
	author={Maracani, Andrea and Michieli, Umberto and Toldo, Marco and Zanuttigh, Pietro},
	booktitle={International Conference on Computer Vision (ICCV)},
	year={2021}}

@inproceedings{NEST,
	title={Early Preparation Pays Off: New Classifier Pre-tuning for Class Incremental Semantic Segmentation},
	author={Xie, Zhengyuan and Lu, Haiquan and Xiao, Jia-wen and Wang, Enguang and Zhang, Le and Liu, Xialei},
	booktitle={European Conference on Computer Vision (ECCV)},
	year={2024}}

@article{CIT,
	title={CIT: Rethinking class-incremental semantic segmentation with a Class Independent Transformation},
	author={Ge, Jinchao and Zhang, Bowen and Liu, Akide and Phan, Vu Minh Hieu and Chen, Qi and Shu, Yangyang and Zhao, Yang},
	journal={Pattern Recognition},
	year={2025}}

@inproceedings{MBS,
	title={Mitigating Background Shift in Class-Incremental Semantic Segmentation},
	author={Park, Gilhan and Moon, WonJun and Lee, SuBeen and Kim, Tae-Young and Heo, Jae-Pil},
	booktitle={European Conference on Computer Vision (ECCV)},
	year={2024}}

@inproceedings{CoLoR,
	title={Continual Learning with Low Rank Adaptation}, 
	author={Martin Wistuba and Prabhu Teja Sivaprasad and Lukas Balles and Giovanni Zappella},
	booktitle={Workshops of Advances in Neural Information Processing (NeurIPS)},
	year={2023}}

@article{taskArithmeticLoRA,
	title={Task Arithmetic with LoRA for Continual Learning},
	author={Rajas Chitale and Ankit Vaidya and Aditya Kane and Archana Ghotkar},
	journal={arXiv:2311.02428},
	year={2023}}

@inproceedings{taskArithmetic,
	title={Editing Models with Task Arithmetic}, 
	author={Gabriel Ilharco and Marco Tulio Ribeiro and Mitchell Wortsman and Suchin Gururangan and Ludwig Schmidt and Hannaneh Hajishirzi and Ali Farhadi},
	booktitle={International Conference on Learning Representations (ICLR)},
	year={2023}}

@inproceedings{LAE,
	title={A Unified Continual Learning Framework with General Parameter-Efficient Tuning}, 
	author={Qiankun Gao and Chen Zhao and Yifan Sun and Teng Xi and Gang Zhang and Bernard Ghanem and Jian Zhang},
	booktitle={International Conference on Computer Vision (ICCV)},
	year={2023}}

@article{OLoRA,
	title={Orthogonal Subspace Learning for Language Model Continual Learning}, 
	author={Xiao Wang and Tianze Chen and Qiming Ge and Han Xia and Rong Bao and Rui Zheng and Qi Zhang and Tao Gui and Xuanjing Huang},
	journal={arXiv:2310.14152},
	year={2023}}

@inproceedings{InfLoRA,
	title={Inflora: Interference-free low-rank adaptation for continual learning},
	author={Liang, Yan-Shuo and Li, Wu-Jun},
	booktitle={Conference on Computer Vision and Pattern Recognition (CVPR)},
	year={2024}}

@article{CLoRA,
      title={CLoRA: Parameter-Efficient Continual Learning with Low-Rank Adaptation}, 
      author={Shishir Muralidhara and Didier Stricker and René Schuster},
      journal={arXiv:2507.19887},
      year={2025}}

@article{peftSurvey,
	title={Parameter-efficient fine-tuning methods for pretrained language models: A critical review and assessment},
	author={Xu, Lingling and Xie, Haoran and Qin, Si-Zhao Joe and Tao, Xiaohui and Wang, Fu Lee},
	journal={arXiv:2312.12148},
	year={2023}}

@inproceedings{adapterDrop,
	title={Adapterdrop: On the efficiency of adapters in transformers},
	author={R{\"u}ckl{\'e}, Andreas and Geigle, Gregor and Glockner, Max and Beck, Tilman and Pfeiffer, Jonas and Reimers, Nils and Gurevych, Iryna},
	booktitle={Conference on Empirical Methods in Natural Language Processing (EMNLP)},
	year={2021}}

@article{IA3,
	title={Few-shot parameter-efficient fine-tuning is better and cheaper than in-context learning},
	author={Liu, Haokun and Tam, Derek and Muqeeth, Mohammed and Mohta, Jay and Huang, Tenghao and Bansal, Mohit and Raffel, Colin A},
	journal={Advances in Neural Information Processing Systems (NeurIPS)},
	year={2022}}

@inproceedings{promptTuning,
	title={The power of scale for parameter-efficient prompt tuning},
	author={Lester, Brian and Al-Rfou, Rami and Constant, Noah},
	booktitle={Conference on Empirical Methods in Natural Language Processing (EMNLP)},
	year={2021}
}

@article{prefixTuning,
	title={Prefix-tuning: Optimizing continuous prompts for generation},
	author={Li, Xiang Lisa and Liang, Percy},
	journal={arXiv:2101.00190},
	year={2021}}

@inproceedings{peftMasking,
	title={Masking as an efficient alternative to finetuning for pretrained language models},
	author={Zhao, Mengjie and Lin, Tao and Mi, Fei and Jaggi, Martin and Sch{\"u}tze, Hinrich},
	booktitle={Conference on Empirical Methods in Natural Language Processing (EMNLP)},
	year={2020}
}

@article{Bitfit,
	title={Bitfit: Simple parameter-efficient fine-tuning for transformer-based masked language-models},
	author={Zaken, Elad Ben and Ravfogel, Shauli and Goldberg, Yoav},
	journal={arXiv:2106.10199},
	year={2021}}

@article{peftPruning,
	title={Parameter-efficient transfer learning with diff pruning},
	author={Guo, Demi and Rush, Alexander M and Kim, Yoon},
	journal={arXiv:2012.0746},
	year={2020}}

@inproceedings{LoRA,
	title={LoRA: Low-Rank Adaptation of Large Language Models},
	author={Hu, Edward J and Wallis, Phillip and Allen-Zhu, Zeyuan and Li, Yuanzhi and Wang, Shean and Wang, Lu and Chen, Weizhu and others},
	booktitle={International Conference on Learning Representations (ICLR)},
	year={2021}}

@article{DyLoRA,
	title={Dylora: Parameter efficient tuning of pre-trained models using dynamic search-free low-rank adaptation},
	author={Valipour, Mojtaba and Rezagholizadeh, Mehdi and Kobyzev, Ivan and Ghodsi, Ali},
	journal={arXiv:2210.07558},
	year={2022}}

@article{cl_pretrained_classification,
  title={Simpler is better: off-the-shelf continual learning through pretrained backbones},
  author={Pelosin, Francesco},
  journal={arXiv:2205.01586},
  year={2022}}

@article{simple_pretrained,
  title={A simple baseline that questions the use of pretrained-models in continual learning},
  author={Janson, Paul and Zhang, Wenxuan and Aljundi, Rahaf and Elhoseiny, Mohamed},
  journal={arXiv:2210.04428},
  year={2022}}

@article{APER,
  title={Revisiting class-incremental learning with pre-trained models: Generalizability and adaptivity are all you need},
  author={Zhou, Da-Wei and Cai, Zi-Wen and Ye, Han-Jia and Zhan, De-Chuan and Liu, Ziwei},
  journal={International Journal of Computer Vision (IJCV)},
  year={2025}}

@article{RanPAC,
  title={Ranpac: Random projections and pre-trained models for continual learning},
  author={McDonnell, Mark D and Gong, Dong and Parvaneh, Amin and Abbasnejad, Ehsan and Van den Hengel, Anton},
  journal={Advances in Neural Information Processing Systems (NeurIPS)},
  year={2023}
}

@inproceedings{SCR,
  title={Supervised contrastive replay: Revisiting the nearest class mean classifier in online class-incremental continual learning},
  author={Mai, Zheda and Li, Ruiwen and Kim, Hyunwoo and Sanner, Scott},
  booktitle={Conference on Computer Vision and Pattern Recognition (CVPR)},
  year={2021}}

@inproceedings{SAM,
  title={Segment anything},
  author={Kirillov, Alexander and Mintun, Eric and Ravi, Nikhila and Mao, Hanzi and Rolland, Chloe and Gustafson, Laura and Xiao, Tete and Whitehead, Spencer and Berg, Alexander C and Lo, Wan-Yen and others},
  booktitle={International Conference on Computer Vision (ICCV)},
  year={2023}}

@article{PASCAL,
	title={The PASCAL Visual Object Classes (VOC) Challenge},
	author={Everingham, Mark and van Gool, Luc and Williams, Christopher KI and Winn, John and Zisserman, Andrew},
	journal={International Journal of Computer Vision (IJCV)},
	year={2010}}

@inproceedings{cityscapes,
	title={The cityscapes dataset for semantic urban scene understanding},
	author={Cordts, Marius and Omran, Mohamed and Ramos, Sebastian and Rehfeld, Timo and Enzweiler, Markus and Benenson, Rodrigo and Franke, Uwe and Roth, Stefan and Schiele, Bernt},
	booktitle={Conference on Computer Vision and Pattern Recognition (CVPR)},
	year={2016}}

@inproceedings{CLIP,
  title={Learning transferable visual models from natural language supervision},
  author={Radford, Alec and Kim, Jong Wook and Hallacy, Chris and Ramesh, Aditya and Goh, Gabriel and Agarwal, Sandhini and Sastry, Girish and Askell, Amanda and Mishkin, Pamela and Clark, Jack and others},
  booktitle={International Conference on Machine Learning (ICML)},
  year={2021}}

@inproceedings{Swin,
  title={Swin Transformer: Hierarchical Vision Transformer using Shifted Windows},
  author={Liu, Ze and Lin, Yutong and Cao, Yue and Hu, Han and Wei, Yixuan and Zhang, Zheng and Lin, Stephen and Guo, Baining},
  booktitle={International Conference on Computer Vision (ICCV)},
  year={2021}}

@article{MicroSeg,
  title={Mining unseen classes via regional objectness: A simple baseline for incremental segmentation},
  author={Zhang, Zekang and Gao, Guangyu and Fang, Zhiyuan and Jiao, Jianbo and Wei, Yunchao},
  journal={Advances in neural information processing systems},
  year={2022}}

@inproceedings{Mask2Former,
  title={Masked-attention mask transformer for universal image segmentation},
  author={Cheng, Bowen and Misra, Ishan and Schwing, Alexander G and Kirillov, Alexander and Girdhar, Rohit},
  booktitle={Conference on Computer Vision and Pattern Recognition},
  year={2022}}

@inproceedings{COCO,
  title={Microsoft coco: Common objects in context},
  author={Lin, Tsung-Yi and Maire, Michael and Belongie, Serge and Hays, James and Perona, Pietro and Ramanan, Deva and Doll{\'a}r, Piotr and Zitnick, C Lawrence},
  booktitle={European Conference on Computer Vision (ECCV)},
  year={2014}}
}

\end{document}